# Face recognition for monitoring operator shift in railways

S Ritika, Dattaraj Rao
General Electric

## ABSTRACT
Train Pilot is a very tedious and stressful job. Pilots must be vigilant at all times and its easy for them to lose track of time of shift. In countries like USA the pilots are mandated by law to adhere to 8 hour shifts. If they exceed 8 hours of shift the railroads may be penalized for over-tiring their drivers. The problem happens when the 8 hour shift may end in middle of a journey. In such case, the new drivers must be moved to the location locomotive is operating for shift change. Hence accurate monitoring of drivers during their shift and making sure the shifts are scheduled correctly is very important for railroads.

Here we propose an automated camera system that uses camera mounted inside Locomotive cabs to continuously record video feeds. These feeds are analyzed in real-time to detect the face of driver and recognize the driver using state-of-the-art deep Learning techniques. The outcome is an increased safety of train pilots. Cameras continuously capture video from inside the cab which is stored on an onboard data acquisition device. Using advanced computer vision and deep learning techniques the videos are analyzed at regular intervals to detect presence of the pilot and identify the pilot. Using a time based analysis, it is identified for how long that pilot's shift has been active. If this time exceeds allocated shift time (typically 8 hours) an alert is sent to the dispatch to adjust shift hours.

## INTRODUCTION
Railroads are vast networks which are always active. They are a pivotal part of economy and any disruption in the flow of rail traffic can cause huge economic loss. Though there has been lot of work on automating rail roads, rail operators still play a very important role in smooth operation of railways. The job of operator is very tedious, involving constant attention on rail road, signals, speed limits whilst ensuring proper operation of locomotive. Operators are bound to work for specified hours and overdue is not allowed according to railroad mandate. But since the train pilot needs to be vigilant throughout the shift, there might be a tendency to lose track. Over straining can be a cause to fatigue which can reduce the efficiency of the operator. Such scenarios should be avoided as they can lead to errors. Operator shift is usually monitored manually or using biometric. Often there isn't a centralized system to monitor shifts.  Also, manual process is prone to manual errors. Thus, there is scope for improving the method to monitor operator shifts.

## PROBLEM STATEMENT
To adhere to the rail road mandate and to ensure safety, there is a need to stick to the Train Pilot is a very tedious job and usually leads to very high stress. The pilot has to be vigilant at all times and its easy for them to lose track of time of shift. Using an automated camera system their shift is monitored and they are prevented from over stretching beyond their shift. This leads to increased safety for Train Pilots. If this time exceeds allocated shift time (typically 8 hours) an alert is sent to the dispatch to adjust shift hours.

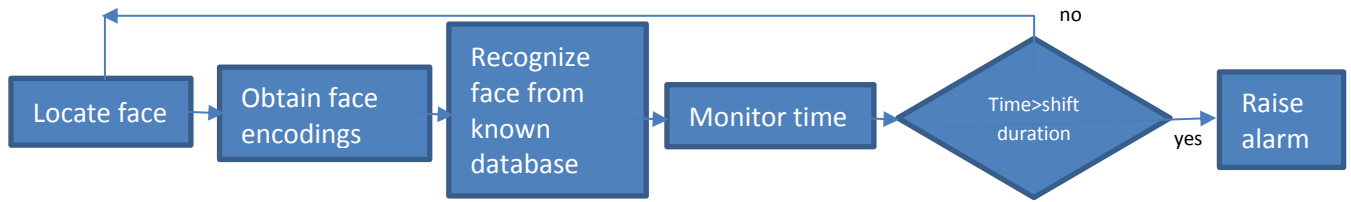

Figure 1 Operator shift monitoring pipeline

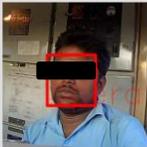

Figure 2 An example of rail operator shift report for a particular day

# APPROACH

To locate the faces in a given image, facial landmark estimation is used. The grey scaled image is converted into histogram of gradients and facial landmarks [1] are identified in the same. This gives the location of the faces in the image as well as 68 facial landmarks for each of them. These landmarks are then used to warp the face and ensure that the facial landmarks are at the same location for all the images. This is done to cater the problem of pose variability. Thus, it makes the problem a bit easier to solve.

Once the face is isolated and warped, the image is fed to a face recognition network [2]. Openface [3] network has been used for this application. It is a Siamese network [4] which has been pretrained to recognize faces. Taking face as input, it outputs 128 encodings [5] for that face. The Siamese network has been trained to minimize the triplet loss, i.e. minimize the norm distance between the encoding of two images of the same person while maximize the distance of images of two different people.

A database should comprise of one image of all the people to be analyzed. With input as video feed of an inward facing camera from the locomotive deck, faces in the video feed can be tracked continuously. If a particular face is recognized for duration more than the operator shift, an alarm can be raised for the same.

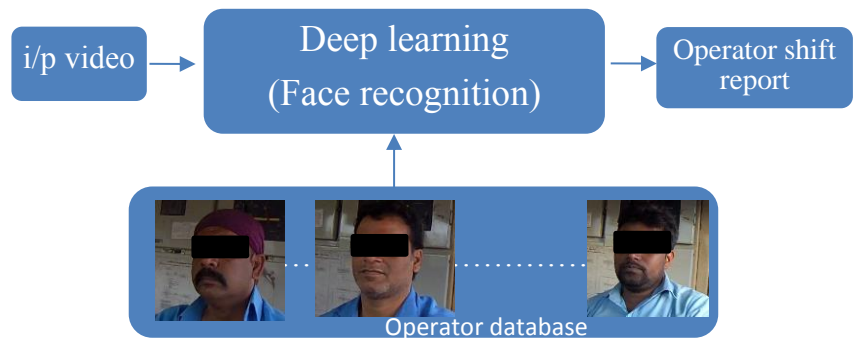

## EXPERIMENTAL RESULTS

The input for this application was an HD video feed at 30 frames per second from an inward facing camera in the locomotive cab. The mounting of camera ensured that it had minimal vibrations as the locomotive moved. One frame was analyzed per 20 seconds. The faces were detected in the frame using HOG facial landmark detection. They were fed to the Siamese network described above. The encodings obtained were compared with a database of the rail pilots comprising one image per person. The recognized faces were tracked with time. If a particular person was observed in the frame for more than the mandated duration, an alert popped on the video. The same logic can be used to highlight any unrecognized face in the operator cab to flag any trespassing efforts. Figure 3 illustrates some outputs from the shift monitoring algorithm.

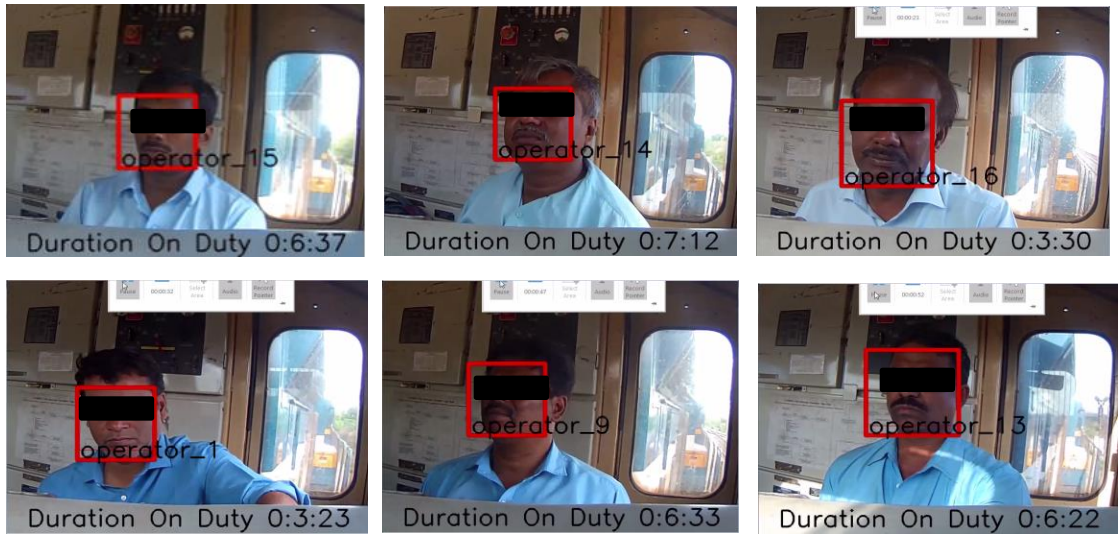

*Figure 3 Snapshots of output of shift monitoring algorithm*

# CONCLUSION AND FUTURE SCOPE

An operator shift monitoring pipeline of detecting face, obtaining face encoding, comparing encoding with the known database of operators and tracking the shift duration was used. The video feed from the camera mounted inside the locomotive cab was taken as input. A database of 16 operators was made for a data spanning a month of videos. The algorithm was seen to track the operators efficiently and come up with the report of the duration of their duty. This approach can be used for real time monitoring of operator shift. The shift report can be sent to the control room real time so that shift timings are maintained correctly. It can also be tuned to raise an alarm if an unknown face person is detected in the operator cab to ensure locomotive security.